\pdfoutput=1
\documentclass[11pt]{article}
\usepackage{times}
\usepackage{latexsym}
\usepackage{graphicx}
\usepackage{amsmath}
\usepackage{booktabs}
\usepackage{multirow}
\usepackage{enumitem}
\usepackage{xcolor}

\definecolor{mygreen}{HTML}{00786C}
\newcommand{\hlc}[2][yellow]{{%
    \colorlet{foo}{#1}%
    \sethlcolor{foo}\hl{#2}}%
}

\usepackage{soul}

\usepackage{ACL2023}

\usepackage{times}
\usepackage{latexsym}

\usepackage[T1]{fontenc}
\usepackage[utf8]{inputenc}
\usepackage{pgfplots}
\DeclareUnicodeCharacter{2212}{−}
\usepgfplotslibrary{groupplots,dateplot}
\usetikzlibrary{patterns,shapes.arrows}
\pgfplotsset{compat=newest}

\usepackage{microtype}

\usepackage{inconsolata}

\title{Concise Answers to Complex Questions:\\Summarization of Long-form Answers}

\author{Abhilash Potluri\thanks{$^{*}$Equal contribution.}~~ Fangyuan Xu$^{*}$~~  Eunsol Choi \\
Department of Computer Science \\
 The University of Texas at Austin \\
 \hspace{0.5em} {\texttt{\{acpotluri, fangyuan, eunsol\}@utexas.edu}} \\}

\begin{document}

\definecolor{lightgreen}{rgb}{0.65, 0.93, 0.65}
\definecolor{lightblue}{rgb}{0.65, 0.85, 0.93}
\definecolor{lightred}{rgb}{0.93, 0.56, 0.56}
\definecolor{lightorange}{rgb}{1.0, 0.8, 0.6}
\definecolor{lgrey}{rgb}{0.85, 0.85, 0.85}

\maketitle
\begin{abstract}

Long-form question answering systems provide rich information by presenting paragraph-level answers, often containing optional background or auxiliary information. While such comprehensive answers are helpful, not all information is \textit{required} to answer the question (e.g. users with domain knowledge do not need an explanation of background). Can we provide a concise version of the answer by summarizing it, while still addressing the question? We conduct a user study on summarized answers generated from state-of-the-art models and our newly proposed extract-and-decontextualize approach. We find a large proportion of long-form answers (over 90\%) in the ELI5 domain can be adequately summarized by at least one system, while complex and implicit answers are challenging to compress. We observe that decontextualization improves the quality of the extractive summary, exemplifying its potential in the summarization task. To promote future work, we provide an extractive summarization dataset covering 1K long-form answers and our user study annotations. Together, we present the first study on summarizing long-form answers, taking a step forward for QA agents that can provide answers at multiple granularities. 

\end{abstract}

\section{Introduction}
Long-form answers~\cite{eli5}, as compared to span-based short answers~\cite{Rajpurkar2016SQuAD1Q}, can provide comprehensive answers to a broader set of questions~\cite{Cao2021ControllableOQ,eli5}. While providing comprehensive information in multiple sentences is helpful, users often prefer short and concise answers to their questions when possible~\cite{decontext}. Today's search engines already present concise answers by highlighting the most relevant parts from the passage excerpts. In this paper, we present the first study on summarizing long-form answers. 

Summarizing long-form answers introduces a new challenge in addition to the faithfulness and fluency challenges of generic summarization which mostly focus on news articles~\cite{CNN, xSum}: the summary output should still provide a reasonable answer to the original question. We take inspiration from a recent study~\cite{discourseStructure} that reports that up to 40\% of sentences in long-form answers contain non-essential information, such as providing background information or examples~\cite{Wang2022ModelingEI}, which demonstrates the potential for compressing long-form answer.

We first aim for an extractive summarization model and collect sentence-level annotations on long-form answers, where annotators identify sentences that address the question directly and can serve as the ``summary''.\footnote{This is a simplified annotation task compared to the original discourse study of ~\citet{discourseStructure}.} We collect a dataset covering 1,134 examples, each consisting of a question, a long-form answer, and a set of summary sentences. To improve the extractive summaries collected, we propose a simple and yet novel summarization approach, \textbf{extract-and-decontextualize}, which first extracts summary sentences and re-writes them to stand-alone~\cite{decontext}. Compared to abstractive summarization models trained on noisy distantly supervised datasets (e.g. CNN/DM~\cite{CNN} and XSum \cite{xSum}) which encourage paraphrasing but also hallucinations~\cite{Kryscinski2019NeuralTS,Cao2018FaithfulTT,Kang2020ImprovedNL}, decontextualization makes minimal edits to the original sentence, preserving its meaning while improving its fluency.

\begin{table*}
\footnotesize
\setlength{\tabcolsep}{3pt}
\begin{center}
\begin{tabular}{p{6.5cm}cp{4cm}cc}
\toprule
\textbf{Input} & \textbf{System}& \textbf{Summarized Answer} & \textbf{Adequacy} & \textbf{Faithful} \\
\midrule
\multirow{3}{*}{\parbox[t][6.5cm]{6.5cm}{\textbf{Q:} Why does car sickness seem to hit the hardest when you look down at your phone, book, etc.?  \newline \textbf{A: }The brain perceived motion because it receives information from the eyes, ears, and muscles. \hlc[lightorange]{When these parts send conflicting information, the brain doesn't know which is right and which is wrong, and this is what causes motion sickness.} An example of this is when reading a book while you are in a moving car. To your eyes, the book is stationary while your inner ear and the rest of your body can feel a sense of motion. This would likely cause car sickness.}} & Abstractive & The brain gets confused when it receives conflicting information about motion from different parts of the body, and this can cause car sickness. & Yes & Yes \\
& Gold Extractive & When these parts send conflicting information, the brain doesn't know which is right and which is wrong, and this is what causes motion sickness. & Partially & Yes \\ 
& Decontext & When \hlc[lightred]{-these parts} \hlc[lightblue]{+the eyes, ears, and muscles} send conflicting information, the brain doesn't know which is right and which is wrong, and this is what causes motion sickness. &  Yes & Yes \\ 
\midrule 
\multirow{2}{*}{\parbox[t][6.5cm]{6.5cm}{\textbf{Q:} How come Obama during his supermajority in both houses wasn't able to pass any legislation he wanted?  \newline \textbf{A: }1) Senators are normally seated in January. […]Al Franken (who would've been \#60) was not seated until July 7.2) Ted Kennedy was dying and had not cast a vote since[…]Note that Sept 24-Feb 4 is about 20 working days, due to recess and holidays.3) \hlc[lightorange]{So, for about 20 working days, the Senate Democrats could have broken a filibuster if you could get every single one of them to agree on something.} […] This did not go well.}}  & Abstractive & The Senate Democrats were unable to pass any legislation during Obama's supermajority due to a lack of 60 votes needed to break a filibuster, due to Al Franken not being seated until July 7 and Ted Kennedy's death in August 2009.  \vspace{0.1cm} & Partially & Yes \\
& Gold Extractive & \multirow{2}{*}{\parbox[t][4cm]{4cm}{So, for about 20 working days, the Senate Democrats could have broken a filibuster if you could get every single one of them to agree on something.}} & \multirow{2}{*}{No} & \multirow{2}{*}{No} \\
& / Decontext \vspace{1cm} & & & \\ 
\bottomrule
\end{tabular} 
\end{center}\vspace{-0.4cm}
\caption{We present two examples of questions, long-form answers, their summarized answers produced by different systems, and human evaluation results ("summary adequacy" and "faithfulness"). We highlight the \hlc[lightorange]{gold extractive summaries} we collected.}\vspace{-0.4cm}
\label{tab:sampleSummary}
\end{table*}

How well do summarization approaches perform in this new domain -- can generated summaries provide fluent, adequate answers to the questions, while preserving the semantics of the original long-form answers? We evaluate fine-tuned abstractive summarization model~\cite{pegasus}, prompted large language model (GPT-3)~\cite{Brown2020LanguageMA}, and our extract-and-decontextualize approach with a user study. Table~\ref{tab:sampleSummary} shows two examples from our user study. We find vanilla extractive approach, even with gold sentences, presents inadequate summaries but decontextualizing them makes them on par with GPT-3 abstractive answers. While none of the systems consistently present high-quality summaries (GPT-3 records a 67\% success rate), most questions (95\%) have at least one system that can generate a valid summary, showing the potential for successful compression of long-form answers. Together, we present the first corpus and study on summarizing long-form answers, opening doors for developing more flexible QA systems which provide answers with varying amounts of information. We release our data, code, and user study templates at \url{https://github.com/acpotluri/lfqa_summary}.
 
\section{Background and Motivation}
The focus of our study is to find a \textbf{concise} answer to a complex question~\cite{eli5}. One way to generate a concise answer is through controllable generation, where the long-form QA model is instructed to generate an answer given a pre-specified length. However, long-form question answering remains challenging, both in terms of modeling~\cite{krishna-etal-2021-hurdles} and reliable evaluation~\cite{Xu23eval}. Existing models often hallucinate~\cite {krishna-etal-2021-hurdles, Liu2023EvaluatingVI} even when paired with relevant evidence documents. Instead of generating a concise answer from scratch, we summarize an \emph{existing} long-form answer, leveraging a large amount of user-written long-form answers often in community-driven QA forums like ELI5 in Reddit. 

How feasible would it be to summarize existing long-form answers? \citet{discourseStructure} conducted an in-depth study on the structure of such long-form answers, assigning one of six functional roles (answer, answer summary, organizational sentence, auxiliary information, and example) to each sentence in long-form answer. The study suggests sentences corresponding to ``answer summary" captures the salient information and ``often suffice by themselves as the answer to the question." Furthermore, they suggest up to 40\% of sentences belongs to roles (e.g., auxiliary information) that are not necessary to answer the question, suggesting summarizing existing answer is viable. We follow their study and collect larger-scale data focusing on the ``answer summary" role to study the summarization of long-form answers.

Summarizing existing answers will support providing a consistent answer set of different granularities, where the users can \emph{expand} condensed answer to see a more detailed version of the same answer. Consistent answers at multiple granularities are harder to enforce with a controllable generation approach. For instance, if we generate a five-sentence answer from the raw evidence set, the five-sentence answer \textbf{can} contain information absent in the ten-sentence answer. 

Lastly, retrieval-augmented long-form QA models~\cite{webgpt} resemble query-focused summarization. Query-focused summarization~\cite{Xu2020QueryFM, kulkarni2020aquamuse} often studies challenging multi-document settings, where the input text is summarized focusing on a particular query, provided at inference time content control. A difference to our setting is that a long-form answer is \textit{written} for the question $q$, presenting already synthesized information tailored for the question.\footnote{This is true for two out of three datasets (ELI5/WebGPT, 82\% of our data) we study. In NQ, the paragraphs are written independently, representing the QFS setting.} 

\section{Extractive Summary for Long-form Answers}\label{sec:data}
We first introduce our annotation task of identifying key sentences for long-form answers, which will be used as an extractive summary. Extractive summaries allow easier data collection and evaluation but can suffer from disfluency and incoherence. Thus, we manually evaluate our collected gold extractive summaries in Section~\ref{sec:validity}.

\subsection{Task}
Given a question $q$ and its long-form answer consisting of $n$ sentences $a_1, a_2, ... a_n$, the model makes a binary decision on whether each sentence $a_i$ should be included in the summary. This setup differs from general summarization in having question $q$ as an additional input.

\subsection{Source Data}
We use long-form answer data, (question, answer) pairs, from prior study~\cite{discourseStructure} which compiled three existing LFQA datasets. \textbf{ELI5} ~\cite{eli5} consists of question answer pairs extracted from the subreddit \textit{Explain Like I’m Five}. \textbf{Natural Questions (NQ)}~\cite{nq}: NQ contains Google search queries as the questions, paired with paragraph-level answers from Wikipedia passages identified by annotators. \textbf{WebGPT}~\cite{webgpt} contains answers written by trained human annotators, with the questions sourced from ELI5. The annotator first searches for related documents using a search engine and then constructs the answers with direct references to those documents. We only take answers that passed their validity annotation, which excludes questions with false presupposition, ill-formed queries, and answers that do not provide valid answers. Their preprocessing step also filters answers with more than 15 sentences or less than 3 sentences.
\subsection{Annotation Task}

Given a question and its long-form answer, annotators select a set of summary sentences containing salient information addressing the question. {The annotator interface and instructions are in the appendix.} As saliency is somewhat subjective, we collect three-way annotations for each example. We recruited crowd workers from Amazon Mechanical Turk. {We recruited workers from English-speaking countries, with at least a 95\% acceptance rate on 1000+ HITs.} Each worker was paid \$0.50 per annotation, translating to an hourly rate of \$15. We recorded reasonable agreement (Fleiss' Kappa 0.53) for the annotations.\footnote{\citet{discourseStructure} hired expert annotators (undergraduate linguistics students), as they required annotators to provide sentence-level labels among six functional roles. The expert annotators reached a similar agreement (0.52 Fleiss' kappa) for the ``summary'' role.}

\begin{table}[]\setlength{\tabcolsep}{5pt}
    \centering
    \small
    \begin{tabular}{lrrrrr}
    \toprule
            & \#  & $|q|$ & $|d|$& $|s|$ & $\frac{|s|}{|d|}$   \\ \midrule
            \multicolumn{4}{l}{\textbf{News dataset}}\\
            CNN/DM & 312k & - & 810 (39.8) & 56 (3.7) & 0.09 \\\midrule
            \multicolumn{5}{l}{\textbf{Query-Focused summarization dataset}}\\
           AQuaMuSe & 5.5k & 9 & 9k (0.4k) & 106 (3.8) & 0.02  \\
    \midrule
   \multicolumn{4}{l}{\textbf{ LFQA datasets} }&\\
    ELI5   & 834 & 16 & 113 (6.5) & 32 (1.6) & 0.33 \\ 
    NQ      & 202 & 10 & 140 (5.3) & 47  (1.5) & 0.36 \\ 
    WebGPT & 98 & 15   & 117 (5.6)& 44 (1.9) & 0.39 \\ 
    All     & 1,134 & 15  & 118 (6.2) & 35 (1.6) & 0.33 \\ \bottomrule
    \end{tabular}
    \caption{Summarization dataset statistics, showing the number of examples (\#), the length of question $q$, document to summarize $d$, and summary $s$. For length, we report the average number of tokens and the average number of sentences in the parenthesis.}
    \label{tab:dataStatistics}
\end{table}
\subsection{Dataset Statistics}

Table~\ref{tab:dataStatistics} contains our collected dataset statistics, comparing it to a popular news summarization dataset~\cite{CNN} and a query-focused summarization dataset, AQuaMuSE~\cite{kulkarni2020aquamuse}. To compute the summary length in our dataset, we randomly choose one of three summary annotations. The average number of sentences chosen as summaries by a single annotator was 1.6 out of 6.2 sentences in long-form answers.
The statistics show that our data handles shorter texts and compress less than existing datasets. On average, long-form answers were compressed to about one-third of their original length, with a slightly higher compression rate for ELI5 answers. This aligns with the prior discourse study~\cite{discourseStructure} which reports ELI5 contains sentences that serve other functional roles (like providing an example) more frequently (23\% compared to 5\% and 8\% in NQ/WebGPT datasets), neither of which are likely to be included in the summary.

\subsection{Automatic Extractive Summarization}\label{subsec:extract}
Having collected a new dataset, we evaluate existing extractive summarization models on it. Is it easy for models to identify key sentences from long-form answers? 

\paragraph{Setting} We aggregate all data from three datasets (ELI5, NQ, WebGPT) and split them into 70\% train, 15\% validation, and 15\% test set. We report classification metrics (precision, recall, $F_1$ scores) with summary sentences being the positive class. For each long-form answer, metrics are computed against each of the three references, with the results from the reference with the maximum $F_1$ score reported. We also report exact-match (EM), whether the model-predicted summary sentence set matches any of the three annotations. The training details and hyperparameters can be found in Appendix~\ref{appendix:trainningdetail}.

\paragraph{PreSumm} We use PreSumm \cite{presumm}, a BERT-based extractive summarization model, which was trained on the CNN/DailyMail~\cite{CNN} dataset. {It encodes the document with pre-trained BERT \cite{bert} and outputs a score for each sentence.} We select a threshold for the score at which it is considered a summary sentence to maximize the F1 score on the validation set. We evaluate both the original model (trained on CNN/DM dataset) and the model fine-tuned on our dataset.

\paragraph{T5} We use a sequence-to-sequence model, T5-large \cite{t5}, to classify whether a sentence belongs to the summary or not. {This was the best performing model for fine-grained role classification of long-form answers in \citet{discourseStructure}.} For question prepending input, the input sequence to the model would be: [$q$ $[1]$ $a_{1}$ $[2]$ $a_{2}$ ... $[n]$ $a_{n}$].
The output sentence would then be of the form: [$[1]$ $r_{1}$ $[2]$ $r_{2}$ ... $[n]$ $r_{n}$],
where $r_i$ was a binary class label whether $i$-th answer sentence $a_i$ belongs to the summary or not.

\paragraph{Results} Table \ref{tab:acc} reports model performances on the test set. The result on the validation set can be found in Table~\ref{tab:valAcc} in the appendix. With in-domain fine-tuning, both models are able to accurately predict which sentences belong to the summary. Fine-tuned T5 model shows a strong performance, though underperforming human, especially in exact match. We also find all trained classifiers benefit from having questions as additional input, signifying that questions provide important signals for content selection. While there is room for improvement, results suggest that predicting key sentence sets is not a major hurdle for state-of-the-art language models. Thus, we use the \textbf{gold} extractive summary for our user study (Section \ref{sec:validity}). 

\begin{table}[]\setlength{\tabcolsep}{5pt}
\small
    \centering
    \small
    \begin{tabular}{lcccc}
    \toprule               & P & R & $F_1$ & EM \%  \\ \midrule
     LEAD-2               & 0.41 & 0.74 & 0.51 & 11.4 \\
     LEAD-3               & 0.46 & 0.83 & 0.56 & 5.3 \\
     PreSumm-cnn (A)          & 0.46 & 0.77 & 0.55 & 11.7  \\
     PreSumm-cnn (Q+A)    &  0.53 & 0.78 & 0.60 & 11.0\\
     PreSumm-cnn+ours (A)    & 0.55 & 0.81 & 0.61 & \textbf{36.0} \\
     PreSumm-cnn+ours (Q+A) & 0.55 & \textbf{0.88} & 0.63 & 30.9 \\ 
     T5-ours (A)       & 0.67 & 0.71 & 0.65 & 20.5 \\
     T5-ours (Q+A)    & \textbf{0.70} & 0.78 & \textbf{0.69} & 25.0 \\ \midrule
     Human$^*$ & 0.77 & 0.79 & 0.77 & 41.3\\
    \bottomrule
    \end{tabular}
    \caption{Binary classification accuracy of extractive summarization models on the test set. }
    \label{tab:acc}
\end{table}
\section{Abstractive Summaries for Long form Answers}\label{sec:model}

While we have gold extractive summaries at hand, they often suffer from disfluencies and factual errors~\cite{Zhang2022ExtractiveIN}. We aim to improve this in two ways, (1) by introducing a decontextualization \cite{decontext} model to edit extractive summaries and (2) by using abstractive summarization models. We explore zero-shot transfer from an abstractive summarization model~\cite{pegasus} and prompting an instruction-tuned large language model~\cite{Brown2020LanguageMA}. We experiment with two types of input sequences: (1) long-form answer only as an input (2) the question followed by a separation token and the long-form answer, whenever applicable. In the latter setting, models sometimes output the question as a part of the summary, which we remove with postprocessing.\footnote{For extractive models, we exclude the question if it is chosen as the summary. For abstractive models, we remove the first sentence of the summary if it has high lexical overlap (over 75\% unigram overlap) with the question (which happened for roughly 38\% of the dataset).}

\subsection{Editing Extractive Summary with Decontextualization}
The disfluencies and lack of coherence of extractive summaries are well-known issues, motivating a flurry of abstractive summarization models~\cite{Rush2015ANA,See2017GetTT}. While abstractive models can provide coherent and fluent summaries, one of their major issues is hallucination~\cite{Kryscinski2019NeuralTS,Cao2018FaithfulTT}. Recent work explores \textbf{extract-and-abstract} approaches~\cite{Hsu2018AUM,Liu2018GeneratingWB,Pilault2020OnEA}, aiming to take the best of both worlds. 
Most of these approaches are fine-tuned on an abstractive summarization dataset. As we don't have an abstractive summary of long-form answers at hand, we opt to use a decontextualization model to re-write the extractive summary.

Decontextualization~\cite{decontext} is a text editing task, which aims to rewrite the target sentence in a document such that the edited target sentence can be interpreted when presented alone while preserving its meaning. While its use cases in QA and text retrieval~\cite{Gao2022RARRRA} have been explored, its use case in summarization has not been explored. Earlier prior work~\cite{Clarke2010DiscourseCF, durrett-etal-2016-learning} have studied discourse constraints for summarization -- that for each pronoun included in the summary, the pronoun's antecedent should be included or the pronoun to be rewritten as a full mention to make summary coherent and clear. Decontextualization is well-suited to prevent these common errors of pronouns/concepts being ``orphaned" in extractive summary.

\paragraph{Method} We use an off-the-shelf decontextualization system from recent work~\cite{Chen2021CanNM},\footnote{\url{https://github.com/jifan-chen/QA-Verification-Via-NLI/.}} which trained a T5 3B model on the original decontextualization dataset \cite{decontext} on Wikipedia text. This model takes the concatenation of the Wikipedia page title and a paragraph with the sentence to be decontextualized as input. For ELI5 and WebGPT answers which lack a page title, we consider the question as the title. 

If the title is $t$ and the answer consists of $k$ sentences [$a_1, a_2, \dots, a_k$] with the $i$-th sentence being the target to be decontextualized, the input will be formatted as: 
$$\text{[CLS] }t \text{ [s] } a_1 \dots a_{i - 1} \text{ [s] } a_{i} \text{ [s] } a_{i + 1} \dots a_{k}\text{[s]}$$ where [CLS] is a start token and [s] is a separator token. The model outputs the sequence: $ \text{[CATEGORY] [SEP] } y$, where the category is one of \texttt{DONE} (if it made edits to the sentence in which case $y$ would be the new sentence), \texttt{Unnecessary} (the sentence does not need an edit, already stand-alone), or \texttt{Infeasible} (the sentence is tricky to be made stand-alone with minimal edits).\footnote{In the case of infeasible and unnecessary cases, $y$ would just be the same as $a_{i}$).} We only apply decontextualization when the first sentence in the extractive summary is \textbf{not} included in the summary set (56\% of examples in the dataset), and only decontextualize the first summary sentence.

\begin{table}[]
    \centering
    \footnotesize
    \begin{tabular}{lr|rrr|r}
    \toprule
 Domain &   Pred          &   \texttt{Un} & \texttt{Inf}  &\texttt{Done}& $\Delta$   \\ \midrule
    Wiki (NQ Short)& human & 12.0 & 20.0 & 68.0 & 23\%  \\ 
    Wiki (NQ Short)&model & 14.7 & 26.3 & 59.0 & 13\% \\ \midrule
    \multicolumn{3}{l}{\emph{LFQA Answers}} \\ \vspace{0.3em}
    Wiki (NQ Long) &model & 66.8 & 13.9 & 19.3 & 28\% \\
    ELI5& model & 49.3 & 34.3 & 16.4 & 34\% \\
    Web-GPT &model & 66.6 & 14.6 & 18.8 & 29\% \\ \bottomrule
    \end{tabular}
    \caption{Decontexutalization output statistics. The second column block represents prediction category distribution, where $\texttt{Un}$ represents unnecessary (no edit is necessary), \texttt{Inf} represents infeasible (stand-alone not feasible), \texttt{Done} represents decontextualization attempted.}
    \label{tab:decontextStats}
\end{table}
\paragraph{Decontexutalization Results} Table~\ref{tab:decontextStats} presents basic statistics of the output from decontextualization model. Somewhat surprisingly, the decontextualization model edited only 17.1\% of input examples, diverging significantly from its training distribution where 60\% of examples are edited. For these edited sentences, we report the length increase ($\Delta$), or the average value of \texttt{(len(decontext)-len(original)) / len(original)}, following the original study. While decontextualization is attempted less frequently when it is decontextualized the length of the sentence increases more substantially. More ELI5 sentences were classified as $\texttt{Infeasible}$. We hypothesize that the sentences in ELI5 could be deemed more challenging because of the narrative nature of Reddit posts. We include sample decontextualization outputs in Table \ref{tab:decontext} in the appendix. 

We manually examine decontextualization outputs from ELI5 and Web-GPT to evaluate their performance on out-of-domain, non-Wikipedia texts. We (the authors of this paper) randomly sample 50 examples where the model has made changes, and 50 examples from the entire set. Out of 50 edits, 42 edits were meaning preserving (without introducing factually incorrect contents), and 44 edits successfully decontextualized the sentence (without unresolved or unclear references). On a randomly sampled set of 50 examples, we evaluate whether the category assigned is correct (infeasible, unnecessary, done), finding 45 examples were assigned the correct category. Overall, we found the zero-shot performance of the decontextualization system on the new domain was surprisingly robust. {Recent work~\cite{Eisenstein2022HonestSF} also showed large language model can perform decontextualization robustly when prompted carefully.} We will evaluate decontextualized summaries with a user study in Section~\ref{sec:validity}.

\subsection{Abstractive Models}
In this section, we explore abstractive models for summarization to improve fluency. 
\paragraph{Pegasus}\cite{pegasus} shows promising performance across diverse summarization benchmarks. We examine a suite of Pegasus fine-tuned on various summarization datasets and chose a model fine-tuned on the CNN/DailyMail as it showed the most promising results upon manual inspection. We do not fine-tune it with our extractive dataset to preserve its abstract nature.

\paragraph{GPT-3} Recent work \cite{goyalzeroshotnews2022} has found that GPT-3 \cite{Brown2020LanguageMA} exhibits strong zero-shot performance on several news summarization benchmarks. Unlike fine-tuned abstractive models, prompted language models would not inherit issues from noisy distant supervision training datasets. Thus, we investigate its ability to perform zero-shot long-form answer summarization. Specifically, we used the \texttt{text-davinci-002 model}.\footnote{We set the max generation length to 512 tokens and temperature to 0. The generations were queried on October 19, 2022.} We explore two settings: with and without length control in the prompt, following prior work \cite{goyalzeroshotnews2022}. The prompt with length control is \texttt{``Q: \{question text\} A: \{answer text\} Summarize the above answer in \{length of gold summary\} sentences''}, and the prompt without length control is \texttt{``Q: \{question text\} A: \{answer text\} Summarize the above answer.''}

\subsection{Automatic Evaluation}\label{subsec:auto_eval}
We first aim to perform an automatic evaluation of abstractive systems, using gold extractive summaries as references. While this would not evaluate fluency, automatic metrics measure the content selection of generated abstractive summaries.
\paragraph{Setting} We use the same data split as in Section~\ref{subsec:extract}, and repeat lead baselines: \texttt{LEAD-2} and \texttt{LEAD-3}.  We use established automatic summarization evaluation metrics ROUGE~\cite{rouge} and BERTScore~\cite{bert-score}.\footnote{We use the \texttt{bert-base-uncased} checkpoint.} As our dataset is 3-way annotated, we report the highest ROUGE-L $F_1$ score among the three reference answers and use the same reference answer to compute BERTScore $F_1$. The \texttt{Human} baseline is computed by choosing one extractive summary annotation at random as the reference and doing a pair-wise computation of ROUGE and BERTScore with the other two annotations for that example.

\paragraph{Results} Table \ref{tab:rougeAndBert} reports model performances on the test set. The results on the development set are in Table~\ref{tab:valRougeAndBert} in the appendix. Similar to other domains, lead baselines show strong performances, outperforming  models trained on out-of-domain data (Pegasus, GPT3). Yet, they are inherently limited, covering only $73\%$ of the summary sentences. We see that the abstractive models show better performance with the BERTScore metric compared to the ROUGE-L metric, potentially due to the ROUGE-L metric punishing for paraphrasing. Having the question in addition to the answer improves the performance of the Pegasus model. Having length control also improves the zero-shot performance of GPT-3, similar to the finding from prior work \cite{Goyal2022SNaCCE}. This is a semi-oracle setting as the model is given the summary length.

\begin{table}
    \centering
    \setlength{\tabcolsep}{4.5pt}
    \footnotesize
    
    \begin{tabular}{llccc}\toprule
                 Model & Input & ROUGE & BERTScore & Length   \\\midrule

     LEAD-2      &    A     & 0.553 & 0.673 & 38.18 (2.00) \\
     LEAD-3      &    A    & \textbf{0.652} & 0.711 & 59.40 (3.00) \\
    
     \midrule 
     Pegasus & A       & 0.569 & 0.749 & 43.03 (2.65) \\
     Pegasus & Q+A   & {0.588} & \textbf{0.759} & 43.36 (2.80) \\
     \midrule
      & A+L & 0.460 & 0.647 & 32.17 (1.71) \\
     \multirow{2}{*}{GPT3} & A & 0.457 & 0.638 & 53.01 (2.84) \\
      & Q+A+L       & {0.497} & {0.670} & \textbf{31.34 (1.63)} \\
      & Q+A & 0.484 & 0.662 & 46.12 (2.20) \\
     \midrule
     
    Human& Q+A & 0.811 & 0.881 & 39.41 (1.93)\\ 
     \bottomrule
    \end{tabular}
    \caption{Automatic evaluation results on the test set. For the ``Input" column, \textsc{A} refers to a long answer while \textsc{Q+A} refers to (question, long answer) as an input to the model and \textsc{L} refers to the length of the gold extractive summary in sentences. For length, we present the number of tokens, with the number of sentences in the parenthesis.}
    \label{tab:rougeAndBert}
\end{table}

\section{Human Evaluation of Summary Answers}\label{sec:validity}
So far we have evaluated summarized answers against the gold extractive summary. Yet, we are aware extractive answers themselves are limited and automatic evaluation of summary is non-trivial. To properly evaluate summarizing long-form answers, we launch a user study evaluating four different types of answer summaries: a gold extractive summary, a gold extractive summary that is decontextualized, an abstract summary from Pegasus, and an abstract summary from GPT3. Can the summarized answer present a useful, concise answer that preserves the original meaning of the long-form answer, without producing incoherent discourse structure (e.g., orphaned anaphora)? 

\begin{table*}[]
    \centering
    \footnotesize
    \setlength{\tabcolsep}{4.5pt}
    \begin{tabular}{l|c|ccc|c|ccc|c}
    \toprule
    \multicolumn{1}{c|}{} & \multicolumn{1}{c|}{Summary } & \multicolumn{3}{c|}{Summary Adequacy}                               & \multicolumn{1}{c|}{Faithfulness} & \multicolumn{3}{c|}{Long-Answer Adequacy} & \multirow{2}{*}{Func}\\ 
    {} &  Fluency (Yes)   & \multicolumn{1}{c}{Yes}   & \multicolumn{1}{c}{Partially} & No    &     (Yes)    & \multicolumn{1}{|c}{Yes}   & \multicolumn{1}{c}{Partially} & No  & \\ \midrule
     Kappa & 0.513 & \multicolumn{3}{c|}{0.368}& 0.506 & \multicolumn{3}{c|}{0.474} & \\\midrule
    $\texttt{Pegasus}$         & {89.7} (91.0)    & {62.5 (63.0)} & {31.4 (31.2)}     & 6.1 (5.8) & {83.2} (82.5)     & {81.5} (82.5) & {17.0} (15.9)     & 1.5 (1.6) & 65.7\\
    $\texttt{GOLD}$                & {85.5} (83.6)    & {61.0} (56.6) & {32.6} (36.6)     & 6.4 (6.9) & {84.0} (83.1)    & {81.7} (81.5) & {16.6} (16.4)    & 1.7 (2.1) & 60.1 \\ 
    $\texttt{GOLD++}$    & {88.6} (93.7)& {66.5} (70.4) & {25.9} (21.7)     & 7.6 (7.9) & {84.4} (84.1)      & {82.5} (82.5) & {16.0} (15.3)    & 1.5 (2.1) & \textbf{67.0} \\
    $\texttt{GPT3}$ & \textbf{{94.1} ({94.1})} & \textbf{{67.8} ({71.4})} & {26.5} ({21.7}) & {5.7} ({6.9}) & \textbf{{85.3} ({85.2})} & {81.9} ({82.0}) & {16.4} ({16.4}) & {1.7} ({1.6}) & \textbf{67.0} \\
    
    \bottomrule 
    \end{tabular}
    \vspace{-0.5em}
    \caption{User study results. The first row shows Fleiss' kappa for each question. The rest of the rows present the percentage of examples in each category, with results on the subset of 63 examples where decontextualization modified the extractive summary presented in parenthesis. The last column presents the percentage of functional short answers, meaning they are adequate, fluent, and meaning-preserving.}
    \label{tab:userstudy}
\end{table*}

\subsection{User Study Design}
We design a two-stage interface to evaluate the summarized answer.{
The exact wording and interface can be found in the appendix (Figures \ref{fig:mturk1}, \ref{fig:mturk2}, \ref{fig:mturkinstruction1}, and \ref{fig:mturkinstruction2}).} First, they are shown the summary answer and the question alone, and then, the original long-form answer will be shown to them. 

\paragraph{Stage 1:} The annotators first measure the quality of the summary answer itself. 

\noindent \textsc{Fluency}{ (choices: Yes/No)}: if the answer is grammatical and fluent. {We do not distinguish coherence and fluency as prior study~\cite{Fabbri2021SummEvalRS} reports that annotators often confuse those two dimensions.} 

\noindent \textsc{{Adequacy}}{ (choices: Yes/Partially/No)}: if the summary adequately answers the original question.

\paragraph{Stage 2:}
The annotators then measure \textit{both} the summary and  original long-form answer.

\noindent \textsc{Faithfulness}{ (choices: Yes/No)}: if the summary accurately captures the main idea of a long-form answer regarding the question.

\noindent \textsc{Long-Answer Adequacy} { (choices: Yes/Partially/No)}: if the long-form answer addresses the question adequately. {This annotation \textit{only} evaluates the original long-form answer, as a control to avoid blaming the summarization system when the long answer itself is not adequate. As we filtered out invalid long answers during pre-processing, most answers should be labeled as adequate.
\subsection{User Study Setting}

\paragraph{Data}
We annotate 175 long-form answers paired with four types of summary: (1) summary generated from our best abstractive model ($\texttt{Pegasus}$), (2) gold extractive summary ($\texttt{GOLD}$), (3) gold extractive summary that is decontextualized with automatic decontextualizer system ($\texttt{GOLD++}$) and (4) $\texttt{GPT-3}$ zero shot summaries with length restriction. {We sample 150 examples at random and additionally sample 25 examples where the decontextualization process made edits to the gold extractive summary.} 

The average length of the tokens for the four summary settings were 43.4, 40.9, 47.6, and 31.3 for $\texttt{Pegasus,GOLD,GOLD++,GPT3}$.

\paragraph{Annotators}
Human evaluation was done on the Amazon Mechanical Turk platform. We required the workers to be from English-speaking countries and have at least a 95\% acceptance rate on 1000+ HITs. Each worker was paid \$0.50 per annotation, translating to an hourly rate of \$15. We set up the task that each annotator will see only one variant of the summary per each long-form answer. The annotators were not aware of which summarization system provided the summary. {A small subset of data is annotated by the authors, following the same setup. We had 561 unique annotators for this task.}

\subsection{Results}
Table~\ref{tab:userstudy} presents the results from the user study. We report two numbers -- one on all 175 examples, and one on a subset of 63 examples where decontextualization changed the extractive summary.

We include the inter-annotator agreement for each question in the first row. We observed moderate to high agreement for all four questions. Evaluating the quality of answers (summary adequacy and long answer adequacy) was more subjective than evaluating fluency or faithfulness, revealing the challenge of open-ended long-form answer evaluation as pointed out in prior work~\cite{krishna-etal-2021-hurdles}. We also see high agreement among annotators by comparing long answer adequacy distributions across four rows, which are very similar as expected.

{Can a summarized version of long-form answers provide an \textbf{adequate} answer to the original question?} We see somewhat mixed results -- while the annotators said the summaries provide at least a partial answer to the question most of the time (over 90\%), only about 60\% of answers per system provide adequate answers. Again, we find that decontextualization helps -- on about 10\% examples, annotators labeled extractive answers as partially adequate, but their decontextualized versions are adequate.\footnote{This difference was also statistically significant with a t-test where Yes/Partially/No maps to a (1.0/0.5/0.0) score.} $\texttt{GPT-3}$ produces adequate summaries the most, showcasing its powerful zero-shot summarization ability~\cite{goyalzeroshotnews2022}. Further analysis showed that summary adequacy is highly system dependent rather than question dependent -- for 90\% of the questions, there is at least one system whose outputs are adequate according to the majority of the annotators. 

\begin{figure*}
    \vspace{-0.8em}
    \footnotesize
    \centering
    \begin{tabular}{p{.97\linewidth} }

\hline
        \multicolumn{1}{l}{\textbf{Summarization error} } \\
        \vspace{0.03em}
        \textbf{Q:} Why do most restaurants sell Pepsi instead of Coke, and yet Coke is seen to be a bigger competitor?  \\
        \textbf{A:} Coke sells way more soda by volume than Pepsi. \hlc[lightorange]{As a response, Pepsi offers its products to restaurants at a reduced cost, which is why many restaurants carry it.} But only up to midscale places -- no nice restaurant serves Pepsi, because Coke has more cachėt, and also you need it for mixed drinks.  Note also that McDonald's, the single biggest restaurant chain in the world, serves Coke.\vspace{0.06em}\\   \hline 
   \multicolumn{1}{l}{\textbf{Complex Answer} } \\
        % \hline 
        \vspace{0.03em}
        \textbf{Q:} How is it that the human brain/body sometimes wakes up seconds before an alarm goes off?! \\
        \textbf{A:} Your body does have internal regulation mechanisms, I'm not a doctor and there are plenty who are who can talk more intelligently about the circadian rhythm of the body etc. The other component is psychological. What's happening is an example of confirmation bias. You've woken up a few times almost on the clock (relative to the total number of days you've ever slept in your life). Though this number is astronomical low, you only remember the times you did wake up on the minute. You bias yourself to count those times and subconsciously ignore the other times and thus you feel as though you have an ability to wake up on time. This also happens when people think that they can catch when people are looking at them. You sometimes do and sometimes don't, but the times you don't are not out of the ordinary so you forget them. Thus you only remember catching them and get a false sense of confirmation.\vspace{0.06em}\\   
        \textbf{GPT-3 summary:} The human brain/body sometimes wakes up seconds before an alarm goes off because of the body's internal regulation mechanisms and the psychological phenomenon of confirmation bias. \\ \hline 
    \end{tabular}
    \vspace{-0.2em}
    \caption{Examples with inadequate summaries: In the first example, the \hlc[lightorange]{highlighted} extractive summaries needs further decontextualization. In the second example, the long-form answer is too complex.}\vspace{-0.7em}
    \label{fig:misleadingSummary}
\end{figure*}
We find \textbf{fluency} is not a major issue, both for extractive and abstractive systems. The large-scale language model (GPT3), in particular, provides the most fluent answers. For the extractive summaries, we see a substantial gain (about 10\% on 63 examples where decontextualization changed the input) in fluency by introducing contextualization. The fluency gap between \texttt{Gold} and \texttt{Gold++} was statistically significant on McNemar's test with $p < 0.05$.

We observe a slightly lower performance on \textbf{faithfulness} across four summary systems compared to fluency. While the weaker abstractive model (Pegasus) ranks slightly lower than the extractive model, GPT-3 somewhat surprisingly outperforms extractive approaches in meaning preservation. This mirrors findings from a recent study~\cite{Zhang2022ExtractiveIN} about how extractive summary can also introduce factual errors. Overall, faithfulness has been extensively studied in summarization literature~\cite{fabbri-etal-2022-qafacteval} but mostly in the news domain.

When can we use summarized answers? In the last column, we report the percentage of summary answers that are fluent, adequate, and faithful to the original long-form answer. Decontextualized answers ($\texttt{GOLD++}$) and $\texttt{GPT-3}$ zero-shot summary achieve more promising results than the other two approaches. Of the 168 long answers considered ``adequate” by a majority of the annotators, 160 (95\%) of them has at least one summary that was considered functional by a majority of the annotators. We examine error cases in the next section.

\subsection{What makes it hard for models to summarize long-form answers?}
As we have identified fluency as a minor issue, we specifically look at 60 examples that satisfy all the following conditions: (1) summary is fluent, (2) summary answer is not fully adequate nor faithful, and (3) long-form answer is adequate. 

We identify a few patterns of why the summary answers fall short: (1) around 10\% of them contain summarization errors (e.g. not properly resolving anaphora or hallucination). (2) for around 60\% of examples, adding a few more sentences to the summary was necessary to provide a coherent answer to the question. This is particularly true in cases where the answers are multifaceted (e.g., providing multiple reasons for some phenomena, and the current summary contains only one of them). We also noticed a few cases where disclaimers (e.g., ``I'm talking about poverty in U.S.") or counterexamples in the long-form answer that were not included in the summary, potentially misleading the readers. (3) some long-form answers (around 25\%) are tricky to summarize without massive rewriting as it is explaining a complex procedure (e.g., why the Obama administration could not pass legislation, see the full example in Table~\ref{tab:sampleSummary}). Figure \ref{fig:misleadingSummary} presents two representative failure cases. Future QA models can actively identify questions that require comprehensive v.s. concise answers.

\section{Related Work}

\paragraph{Query/Aspect-focused summarization} Our task is relevant to query-focused summarization, which studies \textit{controllable} summarization with respect to a query \cite{Xu2020QueryFM, Deng2020MultihopIF,zhu-etal-2020-question, vig-etal-2021-exploring} or aspect~\cite{Angelidis2021ExtractiveOS,Hayashi2021WikiAspAD,Ahuja2022ASPECTNEWSAS, kulkarni2020aquamuse}. 
Recently proposed MASH-QA~\cite{zhu-etal-2020-question} dataset on the medical domain presents a question, context document, and extractive answer sentences. Compared to these works which summarize documents written independently of the question into a summary, we aim to compress long-form answers written with respect to the question. Another line of work ~\cite{fabbri-etal-2022-answersumm,Song2017SummarizingAI} studies generating summaries of \textit{multiple} answers to the same question. Lastly, ~\citet{Deng2019JointLO} looks into the same task formulation of summarizing long-form answers, but their evaluation is limited to distantly supervised data. 

\paragraph{Decontextualization for summarization} \citet{Slobodkin2022ControlledTR} proposes the task of controllable text reduction, which rewrites chosen sentences from a document in a coherent manner using existing summarization datasets. They cover longer documents and involve multiple sentences to be decontextualized whereas we reuse a single-sentence decontextualization model \cite{decontext}.

\section{Conclusion and Future Work}
We present the first study on generating concise answers to complex questions. We  collect an extractive summarization dataset in the new summarization domain of long-form answers to support future research. To address this new task, we deploy diverse summarization models, including zero-shot abstractive summarization models and a new decontextualization postprocessing method, which is applied to extractive summaries. Through our comprehensive user study, we find that around 70\% of the summaries can serve as functional, concise answers to the original questions. Our work shows potential for building QA systems that generate answers at different granularities, as well as using decontextualization to improve the faithfulness and fluency of extractive summaries. Future work can also look into applying controllable generation techniques \cite{yang-klein-2021-fudge, Li-2022-DiffusionLM, Qin2022COLDDE} to {generate} answers with different lengths to generate concise answers.

\section*{Limitations}
Our study is limited in scope, studying only English question-answering data. We also acknowledge that the long-form answers we study are not always factually correct, as they can be outdated~\cite{Zhang2021SituatedQAIE} or incorrect as they are crawled from web forums~\cite{eli5}. 

Further, our user study is limited in its scale, evaluating 175 instances, and does not carefully study potentially diverging interpretations from annotators of different demographics. We also do not extensively explore all summarization models, such as the extract-and-abstract approaches mentioned in related work.

\section*{Ethics Statement}
Our data collection and user study protocols do not collect identifiable private information from annotators.  

The question-answering data we annotated comes from an English online forum and might contain biased information. Our annotation is done by crowd-workers recruited from an online platform. We make use of pre-trained language models to generate abstractive summaries, which could suffer from hallucinating unfactual contents \cite{Kang2020ImprovedNL} and perpetuating bias \cite{field-etal-2021-survey}. Thus, more post-processing steps are required before presenting these contents to users. Our user study shows that our proposed method, extract-and-decontextualize, could be one effective post-processing step to reduce hallucination.

\section*{Acknowledgements}
We thank Tanya Goyal, Jessy Li, Jiacheng Xu, and members of the UT Austin NLP community for their helpful feedback on the draft. We thank Jifan Chen for sharing the decontextualization model with us. We also thank the reviewers and meta-reviewer of the ACL community for helpful comments and feedback on the earlier draft of the paper. Lastly, we would like to thank the crowdworkers for their help with our data annotation and user study. The work is partially supported by a gift from Google Faculty Research Award.
\bibliography{anthology,custom}

\begin{thebibliography}{56}
\expandafter\ifx\csname natexlab\endcsname\relax\def\natexlab#1{#1}\fi

\bibitem[{Ahuja et~al.(2022)Ahuja, Xu, Gupta, Horecka, and
  Durrett}]{Ahuja2022ASPECTNEWSAS}
Ojas Ahuja, Jiacheng Xu, Akshay~Kumar Gupta, Kevin Horecka, and Greg Durrett.
  2022.
\newblock Aspectnews: Aspect-oriented summarization of news documents.
\newblock In \emph{ACL}.

\bibitem[{Angelidis et~al.(2021)Angelidis, Amplayo, Suhara, Wang, and
  Lapata}]{Angelidis2021ExtractiveOS}
Stefanos Angelidis, Reinald~Kim Amplayo, Yoshihiko Suhara, Xiaolan Wang, and
  Mirella Lapata. 2021.
\newblock Extractive opinion summarization in quantized transformer spaces.
\newblock \emph{Transactions of the Association for Computational Linguistics},
  9:277--293.

\bibitem[{Brown et~al.(2020)Brown, Mann, Ryder, Subbiah, Kaplan, Dhariwal,
  Neelakantan, Shyam, Sastry, Askell, Agarwal, Herbert-Voss, Krueger, Henighan,
  Child, Ramesh, Ziegler, Wu, Winter, Hesse, Chen, Sigler, Litwin, Gray, Chess,
  Clark, Berner, McCandlish, Radford, Sutskever, and
  Amodei}]{Brown2020LanguageMA}
Tom~B. Brown, Benjamin Mann, Nick Ryder, Melanie Subbiah, Jared Kaplan,
  Prafulla Dhariwal, Arvind Neelakantan, Pranav Shyam, Girish Sastry, Amanda
  Askell, Sandhini Agarwal, Ariel Herbert-Voss, Gretchen Krueger, T.~J.
  Henighan, Rewon Child, Aditya Ramesh, Daniel~M. Ziegler, Jeff Wu, Clemens
  Winter, Christopher Hesse, Mark Chen, Eric Sigler, Mateusz Litwin, Scott
  Gray, Benjamin Chess, Jack Clark, Christopher Berner, Sam McCandlish, Alec
  Radford, Ilya Sutskever, and Dario Amodei. 2020.
\newblock Language models are few-shot learners.
\newblock \emph{ArXiv}, abs/2005.14165.

\bibitem[{Cao and Wang(2021)}]{Cao2021ControllableOQ}
Shuyang Cao and Lu~Wang. 2021.
\newblock Controllable open-ended question generation with a new question type
  ontology.
\newblock \emph{ArXiv}, abs/2107.00152.

\bibitem[{Cao et~al.(2018)Cao, Wei, Li, and Li}]{Cao2018FaithfulTT}
Ziqiang Cao, Furu Wei, Wenjie Li, and Sujian Li. 2018.
\newblock Faithful to the original: Fact aware neural abstractive
  summarization.
\newblock \emph{ArXiv}, abs/1711.04434.

\bibitem[{Chen et~al.(2021)Chen, Choi, and Durrett}]{Chen2021CanNM}
Jifan Chen, Eunsol Choi, and Greg Durrett. 2021.
\newblock Can nli models verify qa systems' predictions?
\newblock \emph{ArXiv}, abs/2104.08731.

\bibitem[{Choi et~al.(2021)Choi, Palomaki, Lamm, Kwiatkowski, Das, and
  Collins}]{decontext}
Eunsol Choi, Jennimaria Palomaki, Matthew Lamm, Tom Kwiatkowski, Dipanjan Das,
  and Michael Collins. 2021.
\newblock \href {http://arxiv.org/abs/2102.05169} {Decontextualization: Making
  sentences stand-alone}.
\newblock \emph{CoRR}, abs/2102.05169.

\bibitem[{Clarke and Lapata(2010)}]{Clarke2010DiscourseCF}
James Clarke and Mirella Lapata. 2010.
\newblock Discourse constraints for document compression.
\newblock \emph{Computational Linguistics}, 36:411--441.

\bibitem[{Deng et~al.(2019)Deng, Lam, Xie, Chen, Li, Yang, and
  Shen}]{Deng2019JointLO}
Yang Deng, Wai Lam, Yuexiang Xie, Daoyuan Chen, Yaliang Li, Min Yang, and Ying
  Shen. 2019.
\newblock Joint learning of answer selection and answer summary generation in
  community question answering.
\newblock In \emph{AAAI Conference on Artificial Intelligence}.

\bibitem[{Deng et~al.(2020)Deng, Zhang, and Lam}]{Deng2020MultihopIF}
Yang Deng, Wenxuan Zhang, and Wai Lam. 2020.
\newblock Multi-hop inference for question-driven summarization.
\newblock In \emph{Conference on Empirical Methods in Natural Language
  Processing}.

\bibitem[{Devlin et~al.(2018)Devlin, Chang, Lee, and Toutanova}]{bert}
Jacob Devlin, Ming{-}Wei Chang, Kenton Lee, and Kristina Toutanova. 2018.
\newblock \href {http://arxiv.org/abs/1810.04805} {{BERT:} pre-training of deep
  bidirectional transformers for language understanding}.
\newblock \emph{CoRR}, abs/1810.04805.

\bibitem[{Durrett et~al.(2016)Durrett, Berg-Kirkpatrick, and
  Klein}]{durrett-etal-2016-learning}
Greg Durrett, Taylor Berg-Kirkpatrick, and Dan Klein. 2016.
\newblock \href {https://doi.org/10.18653/v1/P16-1188} {Learning-based
  single-document summarization with compression and anaphoricity constraints}.
\newblock In \emph{Proceedings of the 54th Annual Meeting of the Association
  for Computational Linguistics (Volume 1: Long Papers)}, pages 1998--2008,
  Berlin, Germany. Association for Computational Linguistics.

\bibitem[{Eisenstein et~al.(2022)Eisenstein, Andor, Bohnet, Collins, and
  Mimno}]{Eisenstein2022HonestSF}
Jacob Eisenstein, Daniel Andor, Bernd Bohnet, Michael Collins, and David Mimno.
  2022.
\newblock Honest students from untrusted teachers: Learning an interpretable
  question-answering pipeline from a pretrained language model.
\newblock \emph{ArXiv}, abs/2210.02498.

\bibitem[{Fabbri et~al.(2021)Fabbri, Kryscinski, McCann, Socher, and
  Radev}]{Fabbri2021SummEvalRS}
A.~R. Fabbri, Wojciech Kryscinski, Bryan McCann, Richard Socher, and Dragomir
  Radev. 2021.
\newblock Summeval: Re-evaluating summarization evaluation.
\newblock \emph{Transactions of the Association for Computational Linguistics},
  9:391--409.

\bibitem[{Fabbri et~al.(2022{\natexlab{a}})Fabbri, Wu, Liu, and
  Xiong}]{fabbri-etal-2022-qafacteval}
Alexander Fabbri, Chien-Sheng Wu, Wenhao Liu, and Caiming Xiong.
  2022{\natexlab{a}}.
\newblock \href {https://doi.org/10.18653/v1/2022.naacl-main.187}
  {{QAF}act{E}val: Improved {QA}-based factual consistency evaluation for
  summarization}.
\newblock In \emph{Proceedings of the 2022 Conference of the North American
  Chapter of the Association for Computational Linguistics: Human Language
  Technologies}, pages 2587--2601, Seattle, United States. Association for
  Computational Linguistics.

\bibitem[{Fabbri et~al.(2022{\natexlab{b}})Fabbri, Wu, Iyer, Li, and
  Diab}]{fabbri-etal-2022-answersumm}
Alexander Fabbri, Xiaojian Wu, Srini Iyer, Haoran Li, and Mona Diab.
  2022{\natexlab{b}}.
\newblock \href {https://doi.org/10.18653/v1/2022.naacl-main.180}
  {{A}nswer{S}umm: A manually-curated dataset and pipeline for answer
  summarization}.
\newblock In \emph{Proceedings of the 2022 Conference of the North American
  Chapter of the Association for Computational Linguistics: Human Language
  Technologies}, pages 2508--2520, Seattle, United States. Association for
  Computational Linguistics.

\bibitem[{Fan et~al.(2019)Fan, Jernite, Perez, Grangier, Weston, and
  Auli}]{eli5}
Angela Fan, Yacine Jernite, Ethan Perez, David Grangier, Jason Weston, and
  Michael Auli. 2019.
\newblock \href {http://arxiv.org/abs/1907.09190} {{ELI5:} long form question
  answering}.
\newblock \emph{CoRR}, abs/1907.09190.

\bibitem[{Field et~al.(2021)Field, Blodgett, Waseem, and
  Tsvetkov}]{field-etal-2021-survey}
Anjalie Field, Su~Lin Blodgett, Zeerak Waseem, and Yulia Tsvetkov. 2021.
\newblock \href {https://doi.org/10.18653/v1/2021.acl-long.149} {A survey of
  race, racism, and anti-racism in {NLP}}.
\newblock In \emph{Proceedings of the 59th Annual Meeting of the Association
  for Computational Linguistics and the 11th International Joint Conference on
  Natural Language Processing (Volume 1: Long Papers)}, pages 1905--1925,
  Online. Association for Computational Linguistics.

\bibitem[{Gao et~al.(2022)Gao, Dai, Pasupat, Chen, Chaganty, Fan, Zhao, Lao,
  Lee, Juan, and Guu}]{Gao2022RARRRA}
Luyu Gao, Zhuyun Dai, Panupong Pasupat, Anthony Chen, Arun~Tejasvi Chaganty,
  Yicheng Fan, Vincent Zhao, N.~Lao, Hongrae Lee, Da-Cheng Juan, and Kelvin
  Guu. 2022.
\newblock Rarr: Researching and revising what language models say, using
  language models.

\bibitem[{Goyal et~al.(2022{\natexlab{a}})Goyal, Li, and
  Durrett}]{goyalzeroshotnews2022}
Tanya Goyal, Junyi~Jessy Li, and Greg Durrett. 2022{\natexlab{a}}.
\newblock News summarization and evaluation in the era of gpt-3.
\newblock \emph{arXiv preprint arXiv:2209.12356}.

\bibitem[{Goyal et~al.(2022{\natexlab{b}})Goyal, Li, and
  Durrett}]{Goyal2022SNaCCE}
Tanya Goyal, Junyi~Jessy Li, and Greg Durrett. 2022{\natexlab{b}}.
\newblock Snac: Coherence error detection for narrative summarization.
\newblock \emph{ArXiv}, abs/2205.09641.

\bibitem[{Hayashi et~al.(2021)Hayashi, Budania, Wang, Ackerson, Neervannan, and
  Neubig}]{Hayashi2021WikiAspAD}
Hiroaki Hayashi, Prashant Budania, Peng Wang, Chris Ackerson, Raj Neervannan,
  and Graham Neubig. 2021.
\newblock Wikiasp: A dataset for multi-domain aspect-based summarization.
\newblock \emph{Transactions of the Association for Computational Linguistics},
  9:211--225.

\bibitem[{Hsu et~al.(2018)Hsu, Lin, Lee, Min, Tang, and Sun}]{Hsu2018AUM}
Wan~Ting Hsu, Chieh-Kai Lin, Ming-Ying Lee, Kerui Min, Jing Tang, and Min Sun.
  2018.
\newblock A unified model for extractive and abstractive summarization using
  inconsistency loss.
\newblock \emph{ArXiv}, abs/1805.06266.

\bibitem[{Kang and Hashimoto(2020)}]{Kang2020ImprovedNL}
Daniel Kang and Tatsunori~B. Hashimoto. 2020.
\newblock Improved natural language generation via loss truncation.
\newblock In \emph{ACL}.

\bibitem[{Krishna et~al.(2021)Krishna, Roy, and
  Iyyer}]{krishna-etal-2021-hurdles}
Kalpesh Krishna, Aurko Roy, and Mohit Iyyer. 2021.
\newblock \href {https://doi.org/10.18653/v1/2021.naacl-main.393} {Hurdles to
  progress in long-form question answering}.
\newblock In \emph{Proceedings of the 2021 Conference of the North American
  Chapter of the Association for Computational Linguistics: Human Language
  Technologies}, pages 4940--4957, Online. Association for Computational
  Linguistics.

\bibitem[{Kryscinski et~al.(2019)Kryscinski, Keskar, McCann, Xiong, and
  Socher}]{Kryscinski2019NeuralTS}
Wojciech Kryscinski, Nitish~Shirish Keskar, Bryan McCann, Caiming Xiong, and
  Richard Socher. 2019.
\newblock Neural text summarization: A critical evaluation.
\newblock In \emph{EMNLP}.

\bibitem[{Kulkarni et~al.(2020)Kulkarni, Chammas, Zhu, Sha, and
  Ie}]{kulkarni2020aquamuse}
Sayali Kulkarni, Sheide Chammas, Wan Zhu, Fei Sha, and Eugene Ie. 2020.
\newblock Aquamuse: Automatically generating datasets for query-based
  multi-document summarization.
\newblock \emph{arXiv preprint arXiv:2010.12694}.

\bibitem[{Kwiatkowski et~al.(2019)Kwiatkowski, Palomaki, Redfield, Collins,
  Parikh, Alberti, Epstein, Polosukhin, Devlin, Lee, Toutanova, Jones, Kelcey,
  Chang, Dai, Uszkoreit, Le, and Petrov}]{nq}
Tom Kwiatkowski, Jennimaria Palomaki, Olivia Redfield, Michael Collins, Ankur
  Parikh, Chris Alberti, Danielle Epstein, Illia Polosukhin, Jacob Devlin,
  Kenton Lee, Kristina Toutanova, Llion Jones, Matthew Kelcey, Ming-Wei Chang,
  Andrew~M. Dai, Jakob Uszkoreit, Quoc Le, and Slav Petrov. 2019.
\newblock \href {https://doi.org/10.1162/tacl_a_00276} {Natural questions: A
  benchmark for question answering research}.
\newblock \emph{Transactions of the Association for Computational Linguistics},
  7:452--466.

\bibitem[{Li et~al.(2022)Li, Thickstun, Gulrajani, Liang, and
  Hashimoto}]{Li-2022-DiffusionLM}
Xiang~Lisa Li, John Thickstun, Ishaan Gulrajani, Percy Liang, and Tatsunori
  Hashimoto. 2022.
\newblock Diffusion-lm improves controllable text generation.
\newblock \emph{ArXiv}, abs/2205.14217.

\bibitem[{Lin(2004)}]{rouge}
Chin-Yew Lin. 2004.
\newblock \href {https://aclanthology.org/W04-1013} {{ROUGE}: A package for
  automatic evaluation of summaries}.
\newblock In \emph{Text Summarization Branches Out}, pages 74--81, Barcelona,
  Spain. Association for Computational Linguistics.

\bibitem[{Liu et~al.(2023)Liu, Zhang, and Liang}]{Liu2023EvaluatingVI}
Nelson~F. Liu, Tianyi Zhang, and Percy Liang. 2023.
\newblock Evaluating verifiability in generative search engines.
\newblock \emph{ArXiv}, abs/2304.09848.

\bibitem[{Liu et~al.(2018)Liu, Saleh, Pot, Goodrich, Sepassi, Kaiser, and
  Shazeer}]{Liu2018GeneratingWB}
Peter~J. Liu, Mohammad Saleh, Etienne Pot, Ben Goodrich, Ryan Sepassi, Lukasz
  Kaiser, and Noam~M. Shazeer. 2018.
\newblock Generating wikipedia by summarizing long sequences.
\newblock \emph{ArXiv}, abs/1801.10198.

\bibitem[{Liu and Lapata(2019)}]{presumm}
Yang Liu and Mirella Lapata. 2019.
\newblock \href {http://arxiv.org/abs/1908.08345} {Text summarization with
  pretrained encoders}.
\newblock \emph{CoRR}, abs/1908.08345.

\bibitem[{Nakano et~al.(2021)Nakano, Hilton, Balaji, Wu, Ouyang, Kim, Hesse,
  Jain, Kosaraju, Saunders et~al.}]{webgpt}
Reiichiro Nakano, Jacob Hilton, Suchir Balaji, Jeff Wu, Long Ouyang, Christina
  Kim, Christopher Hesse, Shantanu Jain, Vineet Kosaraju, William Saunders,
  et~al. 2021.
\newblock \href {https://arxiv.org/abs/2112.09332} {Webgpt: Browser-assisted
  question-answering with human feedback}.
\newblock \emph{arXiv preprint arXiv:2112.09332}.

\bibitem[{Nallapati et~al.(2016)Nallapati, Xiang, and Zhou}]{CNN}
Ramesh Nallapati, Bing Xiang, and Bowen Zhou. 2016.
\newblock \href {http://arxiv.org/abs/1602.06023} {Sequence-to-sequence rnns
  for text summarization}.
\newblock \emph{CoRR}, abs/1602.06023.

\bibitem[{Narayan et~al.(2018)Narayan, Cohen, and Lapata}]{xSum}
Shashi Narayan, Shay~B. Cohen, and Mirella Lapata. 2018.
\newblock \href {http://arxiv.org/abs/1808.08745} {Don't give me the details,
  just the summary! topic-aware convolutional neural networks for extreme
  summarization}.
\newblock \emph{CoRR}, abs/1808.08745.

\bibitem[{Pilault et~al.(2020)Pilault, Li, Subramanian, and
  Pal}]{Pilault2020OnEA}
Jonathan Pilault, Raymond Li, Sandeep Subramanian, and Christopher~Joseph Pal.
  2020.
\newblock On extractive and abstractive neural document summarization with
  transformer language models.
\newblock \emph{Proceedings of the 2020 Conference on Empirical Methods in
  Natural Language Processing (EMNLP)}, page 9308–9319.

\bibitem[{Qin et~al.(2022)Qin, Welleck, Khashabi, and Choi}]{Qin2022COLDDE}
Lianhui Qin, Sean Welleck, Daniel Khashabi, and Yejin Choi. 2022.
\newblock Cold decoding: Energy-based constrained text generation with langevin
  dynamics.
\newblock \emph{ArXiv}, abs/2202.11705.

\bibitem[{Raffel et~al.(2019)Raffel, Shazeer, Roberts, Lee, Narang, Matena,
  Zhou, Li, and Liu}]{t5}
Colin Raffel, Noam Shazeer, Adam Roberts, Katherine Lee, Sharan Narang, Michael
  Matena, Yanqi Zhou, Wei Li, and Peter~J. Liu. 2019.
\newblock \href {http://arxiv.org/abs/1910.10683} {Exploring the limits of
  transfer learning with a unified text-to-text transformer}.
\newblock \emph{CoRR}, abs/1910.10683.

\bibitem[{Rajpurkar et~al.(2016)Rajpurkar, Zhang, Lopyrev, and
  Liang}]{Rajpurkar2016SQuAD1Q}
Pranav Rajpurkar, Jian Zhang, Konstantin Lopyrev, and Percy Liang. 2016.
\newblock Squad: 100,000+ questions for machine comprehension of text.
\newblock In \emph{EMNLP}.

\bibitem[{Rush et~al.(2015)Rush, Chopra, and Weston}]{Rush2015ANA}
Alexander~M. Rush, Sumit Chopra, and Jason Weston. 2015.
\newblock A neural attention model for abstractive sentence summarization.
\newblock In \emph{EMNLP}.

\bibitem[{See et~al.(2017)See, Liu, and Manning}]{See2017GetTT}
A.~See, Peter~J. Liu, and Christopher~D. Manning. 2017.
\newblock Get to the point: Summarization with pointer-generator networks.
\newblock \emph{ArXiv}, abs/1704.04368.

\bibitem[{Slobodkin et~al.(2022)Slobodkin, Roit, Hirsch, Ernst, and
  Dagan}]{Slobodkin2022ControlledTR}
Aviv Slobodkin, Paul Roit, Eran Hirsch, Ori Ernst, and Ido Dagan. 2022.
\newblock Controlled text reduction.

\bibitem[{Song et~al.(2017)Song, Ren, Liang, Li, Ma, and
  de~Rijke}]{Song2017SummarizingAI}
Hongya Song, Zhaochun Ren, Shangsong Liang, Piji Li, Jun Ma, and M.~de~Rijke.
  2017.
\newblock Summarizing answers in non-factoid community question-answering.
\newblock \emph{Proceedings of the Tenth ACM International Conference on Web
  Search and Data Mining}.

\bibitem[{Vig et~al.(2021)Vig, Fabbri, Kry{\'s}ci{\'n}ski, Wu, and
  Liu}]{vig-etal-2021-exploring}
Jesse Vig, Alexander~R. Fabbri, Wojciech Kry{\'s}ci{\'n}ski, Chien-Sheng Wu,
  and Wenhao Liu. 2021.
\newblock \href {http://arxiv.org/abs/2112.07637} {Exploring neural models for
  query-focused summarization}.

\bibitem[{Wang et~al.(2022)Wang, Xu, Thompson, Choi, and
  Iyyer}]{Wang2022ModelingEI}
Shufan Wang, Fangyuan Xu, Laure Thompson, Eunsol Choi, and Mohit Iyyer. 2022.
\newblock Modeling exemplification in long-form question answering via
  retrieval.
\newblock In \emph{North American Chapter of the Association for Computational
  Linguistics}.

\bibitem[{Wolf et~al.(2019)Wolf, Debut, Sanh, Chaumond, Delangue, Moi, Cistac,
  Rault, Louf, Funtowicz, and Brew}]{Wolf2019HuggingFacesTS}
Thomas Wolf, Lysandre Debut, Victor Sanh, Julien Chaumond, Clement Delangue,
  Anthony Moi, Pierric Cistac, Tim Rault, R{\'e}mi Louf, Morgan Funtowicz, and
  Jamie Brew. 2019.
\newblock Huggingface's transformers: State-of-the-art natural language
  processing.
\newblock \emph{ArXiv}, abs/1910.03771.

\bibitem[{Xu et~al.(2022)Xu, Li, and Choi}]{discourseStructure}
Fangyuan Xu, Junyi~Jessy Li, and Eunsol Choi. 2022.
\newblock \href {https://aclanthology.org/2022.acl-long.249/} {How do we answer
  complex questions: Discourse structure of long-form answers}.
\newblock In \emph{Proceedings of the Annual Meeting of the Association for
  Computational Linguistics}.

\bibitem[{Xu et~al.(2023)Xu, Song, Iyyer, and Choi}]{Xu23eval}
Fangyuan Xu, Yixiao Song, Mohit Iyyer, and Eunsol Choi. 2023.
\newblock A critical evaluation of evaluations for long-form question
  answering.

\bibitem[{Xu and Lapata(2020)}]{Xu2020QueryFM}
Yumo Xu and Mirella Lapata. 2020.
\newblock Query focused multi-document summarization with distant supervision.
\newblock \emph{ArXiv}, abs/2004.03027.

\bibitem[{Yang and Klein(2021)}]{yang-klein-2021-fudge}
Kevin Yang and Dan Klein. 2021.
\newblock \href {https://doi.org/10.18653/v1/2021.naacl-main.276} {{FUDGE}:
  Controlled text generation with future discriminators}.
\newblock In \emph{Proceedings of the 2021 Conference of the North American
  Chapter of the Association for Computational Linguistics: Human Language
  Technologies}, pages 3511--3535, Online. Association for Computational
  Linguistics.

\bibitem[{Zhang et~al.(2019)Zhang, Zhao, Saleh, and Liu}]{pegasus}
Jingqing Zhang, Yao Zhao, Mohammad Saleh, and Peter~J. Liu. 2019.
\newblock \href {http://arxiv.org/abs/1912.08777} {{PEGASUS:} pre-training with
  extracted gap-sentences for abstractive summarization}.
\newblock \emph{CoRR}, abs/1912.08777.

\bibitem[{Zhang and Choi(2021)}]{Zhang2021SituatedQAIE}
Michael~J.Q. Zhang and Eunsol Choi. 2021.
\newblock Situatedqa: Incorporating extra-linguistic contexts into qa.
\newblock \emph{ArXiv}, abs/2109.06157.

\bibitem[{Zhang et~al.(2022)Zhang, Wan, and Bansal}]{Zhang2022ExtractiveIN}
Shiyue Zhang, David Wan, and Mohit Bansal. 2022.
\newblock Extractive is not faithful: An investigation of broad unfaithfulness
  problems in extractive summarization.
\newblock \emph{ArXiv}, abs/2209.03549.

\bibitem[{Zhang* et~al.(2020)Zhang*, Kishore*, Wu*, Weinberger, and
  Artzi}]{bert-score}
Tianyi Zhang*, Varsha Kishore*, Felix Wu*, Kilian~Q. Weinberger, and Yoav
  Artzi. 2020.
\newblock \href {https://openreview.net/forum?id=SkeHuCVFDr} {Bertscore:
  Evaluating text generation with bert}.
\newblock In \emph{International Conference on Learning Representations}.

\bibitem[{Zhu et~al.(2020)Zhu, Ahuja, Juan, Wei, and
  Reddy}]{zhu-etal-2020-question}
Ming Zhu, Aman Ahuja, Da-Cheng Juan, Wei Wei, and Chandan~K. Reddy. 2020.
\newblock \href {https://doi.org/10.18653/v1/2020.findings-emnlp.342} {Question
  answering with long multiple-span answers}.
\newblock In \emph{Findings of the Association for Computational Linguistics:
  EMNLP 2020}, pages 3840--3849, Online. Association for Computational
  Linguistics.

\end{thebibliography}
\bibliographystyle{acl_natbib}

\appendix

\section{Appendix}

\subsection{Summary Annotation Interface}

Figure \ref{fig:summary_annotation_interface} presents the interface for summary annotation \ref{sec:data} and Figure \ref{fig:summary_annotation_instruction} is the screenshot of the instruction presented to the annotators.

\subsection{User Study Interface}
\label{sec:appendix}
Figures \ref{fig:mturk1} and \ref{fig:mturk2} are screenshots of the interface provided to the MTurkers who participated in the user study to analyze the quality of the summaries and Figures \ref{fig:mturkinstruction1} and \ref{fig:mturkinstruction2} are screenshots of the instructions provided with the corresponding steps.

\subsection{Dataset Compression Statistics}

Figure \ref{fig:compression} plots the token-level compression ratio (\% of tokens included in the summary) on the three different types of long-form answers we study.

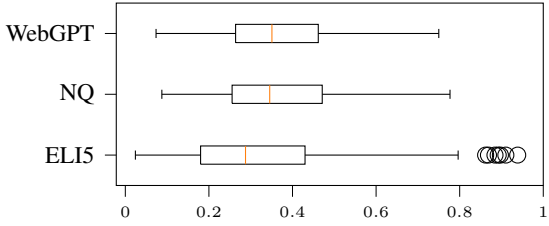
\begin{figure}
    \centering
    % This file was created with tikzplotlib v0.10.1.
\begin{tikzpicture}

\definecolor{darkgray176}{RGB}{176,176,176}
\definecolor{darkorange25512714}{RGB}{255,127,14}

\begin{axis}[
tick align=outside,
tick pos=left,
height=.25\textwidth,
width=.45\textwidth,
x grid style={darkgray176},
xmin=-0.021637570690927, xmax=1,
xtick style={color=black},
xticklabel style={font=\tiny},
y grid style={darkgray176},
ymin=0.5, ymax=3.5,
ytick style={color=black},
ytick={1,2,3},
yticklabel style={font=\small},
yticklabels={ELI5,NQ,WebGPT},
]
\addplot [black]
table {%
0.180094765178799 0.85
0.180094765178799 1.15
0.429994205417934 1.15
0.429994205417934 0.85
0.180094765178799 0.85
};
\addplot [black]
table {%
0.180094765178799 1
0.0240963855421687 1
};
\addplot [black]
table {%
0.429994205417934 1
0.796296296296296 1
};
\addplot [black]
table {%
0.0240963855421687 0.925
0.0240963855421687 1.075
};
\addplot [black]
table {%
0.796296296296296 0.925
0.796296296296296 1.075
};
\addplot [black, mark=o, mark size=3, mark options={solid,fill opacity=0}, only marks]
table {%
0.862068965517241 1
0.91044776119403 1
0.938775510204082 1
0.892156862745098 1
0.884444444444444 1
0.897435897435897 1
0.868852459016393 1
};
\addplot [black]
table {%
0.255486542443064 1.85
0.255486542443064 2.15
0.47114014441936 2.15
0.47114014441936 1.85
0.255486542443064 1.85
};
\addplot [black]
table {%
0.255486542443064 2
0.0874125874125874 2
};
\addplot [black]
table {%
0.47114014441936 2
0.776859504132231 2
};
\addplot [black]
table {%
0.0874125874125874 1.925
0.0874125874125874 2.075
};
\addplot [black]
table {%
0.776859504132231 1.925
0.776859504132231 2.075
};
\addplot [black]
table {%
0.263857333913408 2.85
0.263857333913408 3.15
0.461662946428571 3.15
0.461662946428571 2.85
0.263857333913408 2.85
};
\addplot [black]
table {%
0.263857333913408 3
0.0735294117647059 3
};
\addplot [black]
table {%
0.461662946428571 3
0.75 3
};
\addplot [black]
table {%
0.0735294117647059 2.925
0.0735294117647059 3.075
};
\addplot [black]
table {%
0.75 2.925
0.75 3.075
};
\addplot [darkorange25512714]
table {%
0.287586315128688 0.85
0.287586315128688 1.15
};
\addplot [darkorange25512714]
table {%
0.345490716180371 1.85
0.345490716180371 2.15
};
\addplot [darkorange25512714]
table {%
0.350774352294109 2.85
0.350774352294109 3.15
};
\end{axis}

\end{tikzpicture}
    \caption{Box plot of compression ratio $\frac{|s|}{|d|}$.}
    \label{fig:compression}
\end{figure}

\section{Model Training Details}\label{appendix:trainningdetail}

All models are trained/evaluated on NVIDIA Quadro RTX 8000 GPUs. We use \verb|pytorch-transformers| \citet{Wolf2019HuggingFacesTS} to implement our models. The hyperparameters are manually searched by the authors.

\paragraph{PreSumm}
We use the checkpoint of \texttt{BertSumExt} from \url{https://github.com/nlpyang/PreSumm}. We use the same hyperparameter in the original paper, using a batch size of $16$ and a learning rate of $2e-3$. On two GPUs, fine-tuning on the training set and then evaluating on the test set takes between 1 to 2 hours.

\paragraph{T5}
We use the T5-large checkpoint with 770 million parameters and fine-tune for $30$ epochs with a batch size of $16$ and learning rate of $1e-4$. On two GPUs, fine-tuning on the training set and then evaluating on the test set takes between 2 to 3 hours.

\subsection{Validation Set Results}
Tables \ref{tab:valRougeAndBert} and \ref{tab:valAcc} show our automatic evaluation results on the validation set for the extractive and abstractive models (computed in the same way that the test set values were).

\subsection{Decontextualization Sample Output}\label{appendix:decontext}
Table \ref{tab:decontext} gives three examples of the modifications that the decontextualization models made to the extractive gold label summaries.

\begin{figure*}
    \centering
    \includegraphics[scale=0.36]{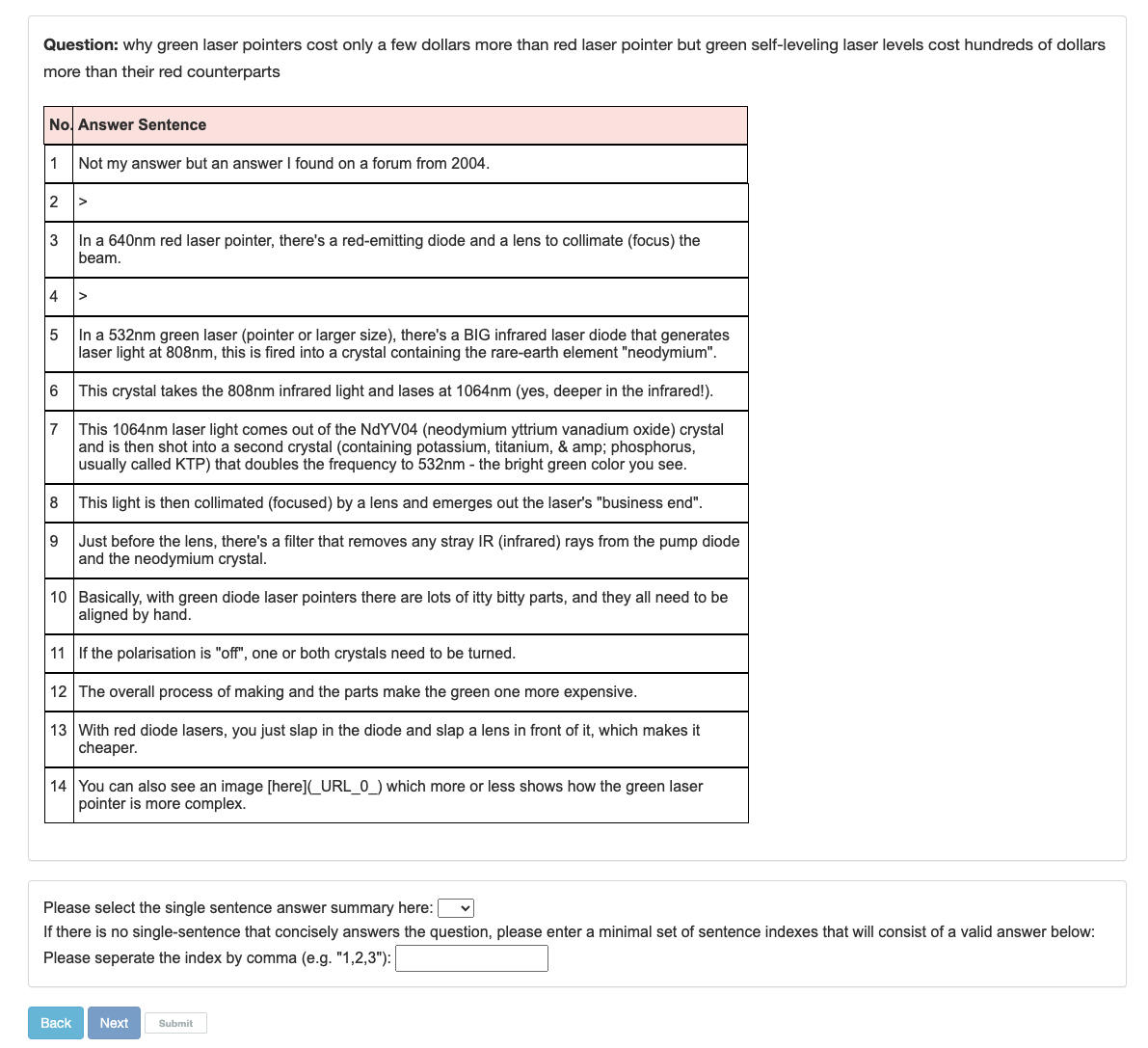}
    \caption{Summary annotation interface}
    \label{fig:summary_annotation_interface}
\end{figure*}

\begin{figure*}
    \centering
    \includegraphics[scale=0.36]{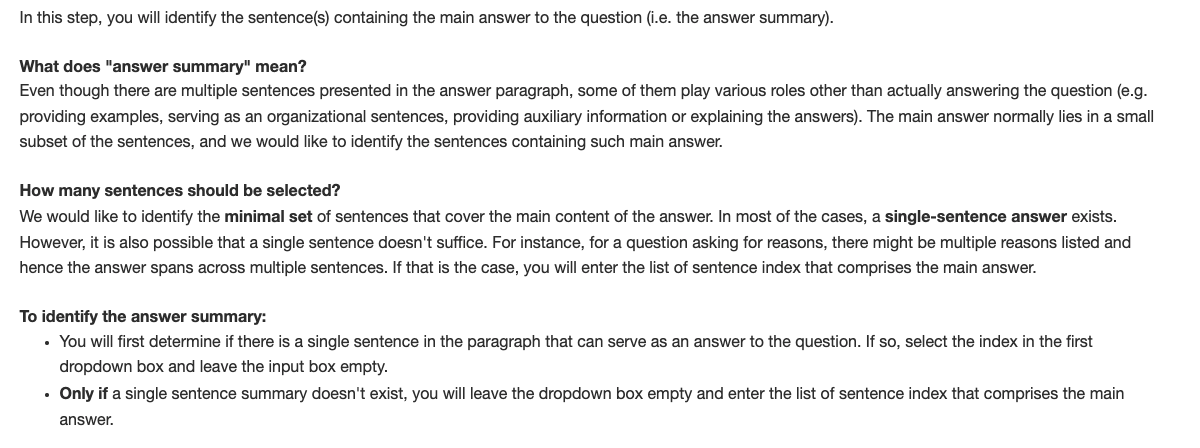}
    \caption{Summary annotation instruction. We provided a few examples to the annotators, which are truncated here.}
    \label{fig:summary_annotation_instruction}
\end{figure*}

\begin{figure*}
    \centering
    \includegraphics[scale=0.48]{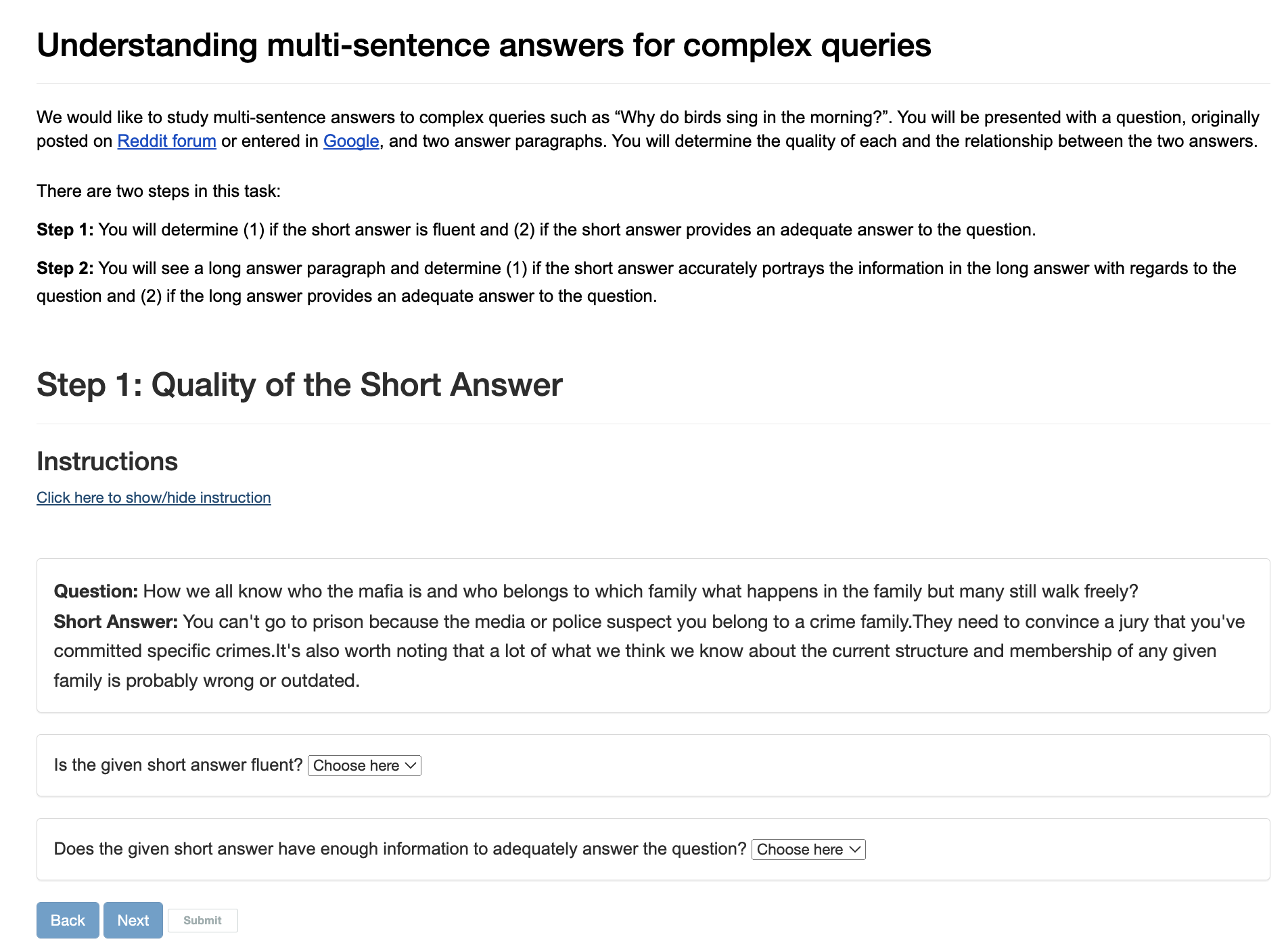}
    \caption{User study annotation UI (Step 1)}
    \label{fig:mturk1}
\end{figure*}

\begin{figure*}
    \centering
    \includegraphics[scale=0.46]{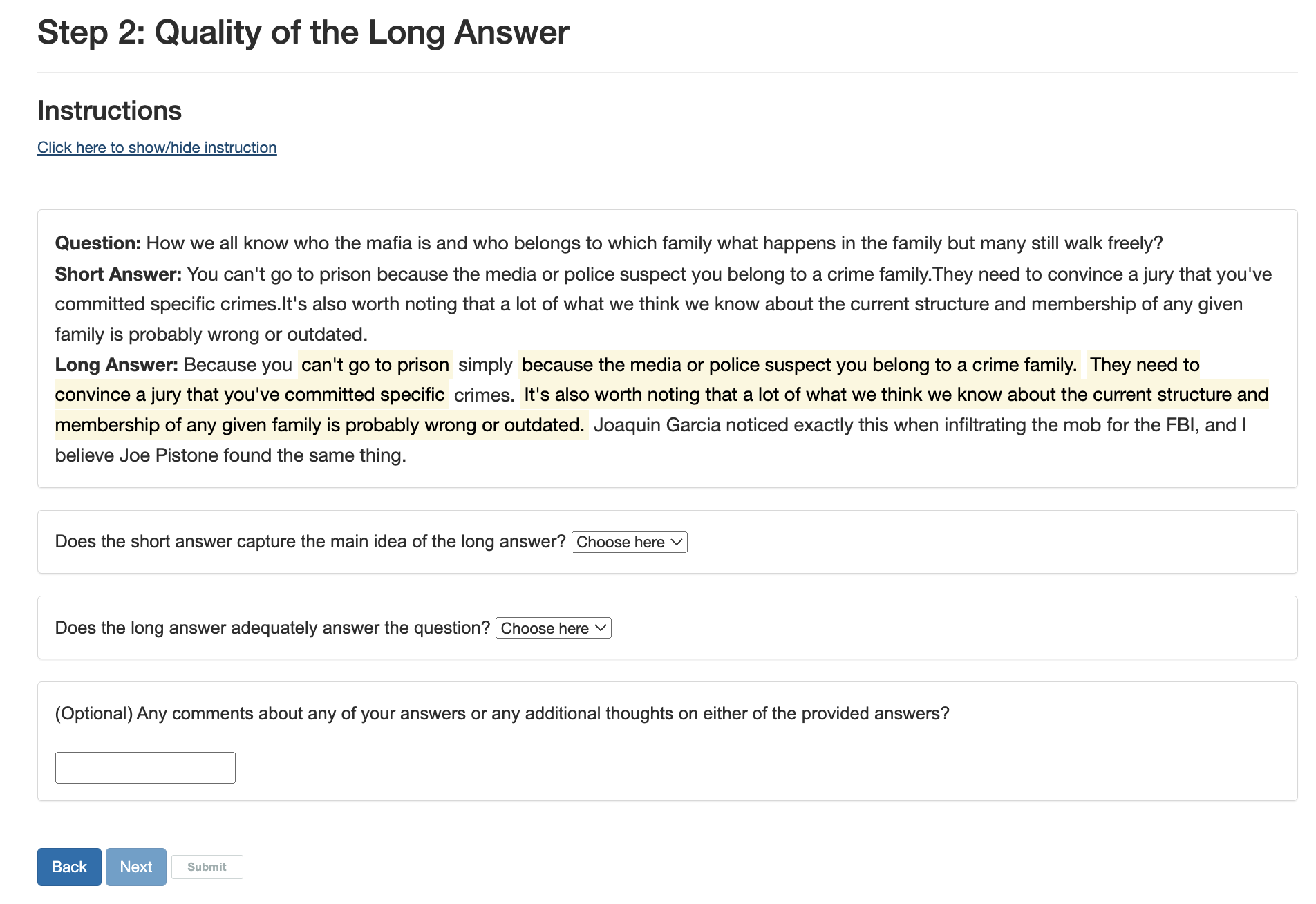}
    \caption{User study annotation UI (Step 2)}
    \label{fig:mturk2}
\end{figure*}

\begin{figure*}
    \centering
    \includegraphics[scale=0.45]{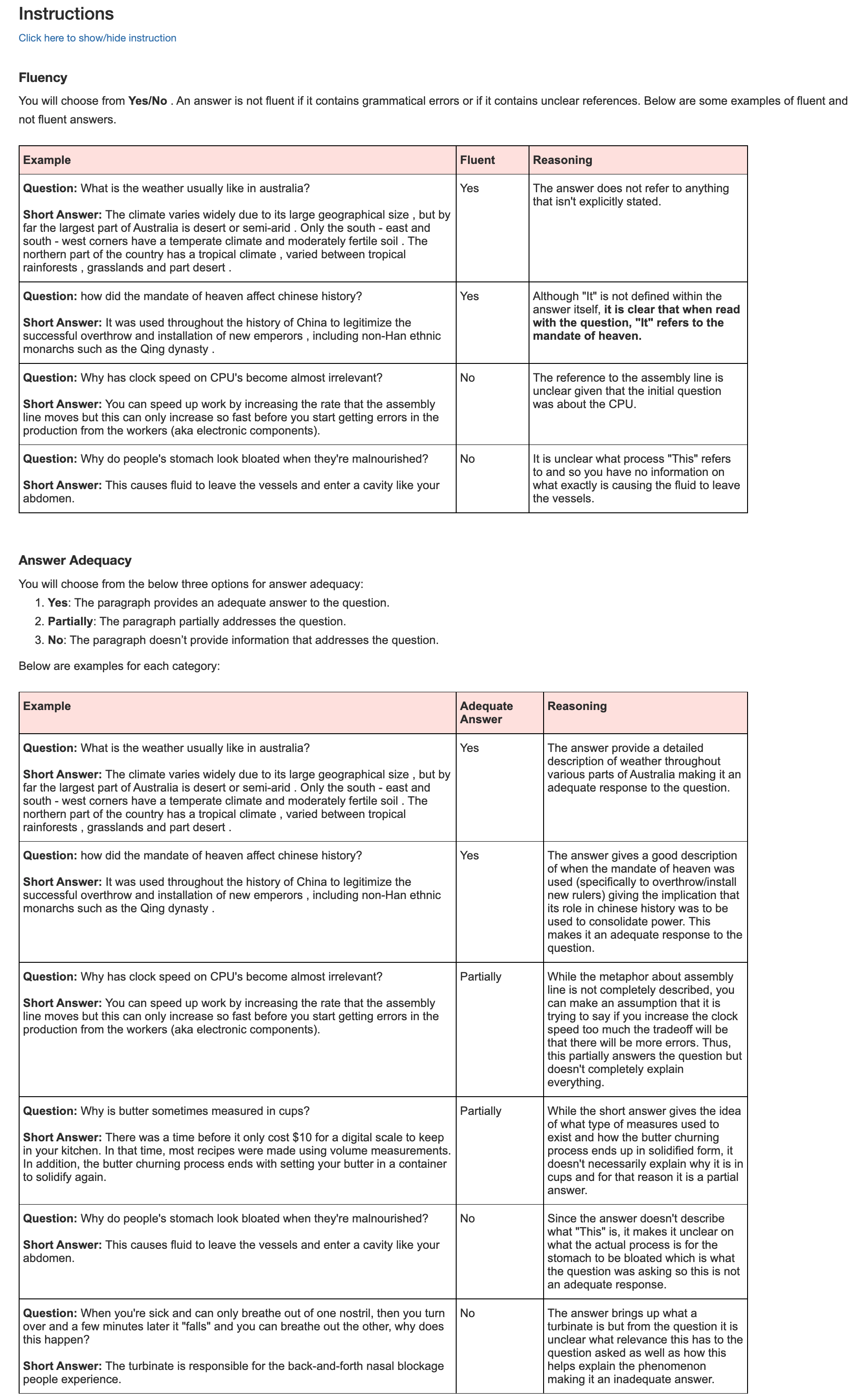}
    \caption{User study instructions (Step 1)}
    \label{fig:mturkinstruction1}
\end{figure*}

\begin{figure*}
    \centering
    \includegraphics[scale=0.46]{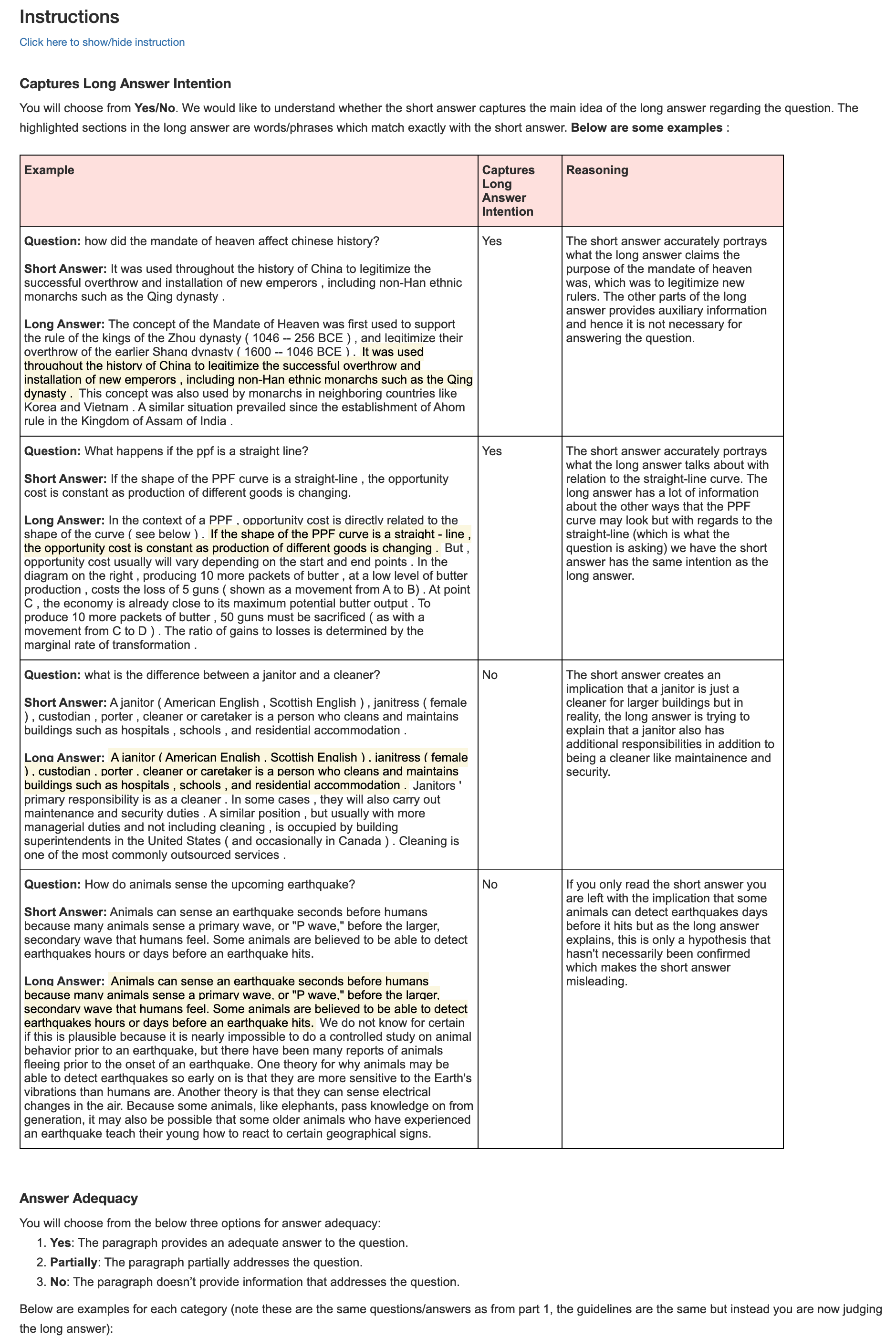}
    \caption{User study instructions (Step 2)}
    \label{fig:mturkinstruction2}
\end{figure*}

\begin{table}[]
    \centering
    \setlength{\tabcolsep}{4.5pt}
    \small
    \begin{tabular}{llccc}\toprule
                 Model & Input  & ROUGE & BERTScore & Length   \\\midrule
    
     LEAD-2      &    A     & 0.541 & 0.677 & 38.36 (2.00) \\
     LEAD-3      &    A     & 0.641 & 0.710 & 58.29 (3.00) \\
     
     \midrule
     Pegasus & A       & 0.571 & 0.750 & 43.50 (2.77) \\
     Pegasus& Q + A   & \textbf{0.572} & \textbf{0.752} & 41.11 (2.64) \\
     \midrule
     GPT3 & (length) & {0.517} & {0.681} & \textbf{32.29 (1.68)} \\
     GPT3 &  & 0.507 & 0.683 & 48.03 (2.24) \\
     \midrule
     Human &  & 0.815 & 0.883 & 39.62 (1.95)\\ 
     \bottomrule
    \end{tabular}
    \caption{Automatic summary evaluation results on the validation set. For the ``Input" column, \textsc{A} refers to using only the long answer as an input to the model while \textsc{Q+A} provides the long answer with the question prepended as an input. The length is computed in the number of tokens/words and the number in parenthesis represents the average number of sentences.}
    \label{tab:valRougeAndBert}
\end{table}

\begin{table}[]
\small
    \centering
    \setlength{\tabcolsep}{5pt}
    \small
    \begin{tabular}{lcccc}
    \toprule               & P & R & $F_1$ & EM \%  \\ \midrule
     LEAD-2               & 0.42 & 0.74 & 0.51 & 11.3 \\
     LEAD-3               & 0.47 & 0.81 & 0.55 & 5.6 \\
     PreSumm-cnn (A)      & 0.47 & 0.75 & 0.55 & 11.7 \\
     PreSumm-cnn (Q+A)    & 0.52 & 0.78 & 0.60 & 21.8 \\
     PreSumm-cnn+ours (A)   & 0.56  & 0.89 & 0.65 & 28.1 \\
     PreSumm-cnn+ours (Q+A) & 0.58 & \textbf{0.91} & 0.68 & 35.9 \\ 
     T5-ours (A)      & 0.70 & 0.73 & 0.66 & 20.0   \\
     T5-ours (Q+A)    & \textbf{0.73} & 0.78 & \textbf{0.71} & 26.3 \\ \midrule
     Human$^*$ & 0.76 & 0.80 & 0.77 & 40.8\\
    \bottomrule
    \end{tabular}
    \caption{Binary classification accuracy of extractive summarization models on the validation set.}
    \label{tab:valAcc}
\end{table}

\begin{table*}[]
    \scriptsize
    \centering
    \begin{tabular}{|p{0.2\linewidth} | p{0.4\linewidth} | p{0.3\linewidth} |}
        \hline Question\rule{0pt}{2.6ex} & Long Answer (Abridged) & Decontextualized Extractive Summary \\
         \hline \rule{0pt}{2.6ex}How did Switzerland stay out of WWII? & They were literally the bankers of the war. \hlc[lgrey]{The Nazis and the allies both kept their assets there.} This is how they stayed neutral, because if either side invaded, that side's assets would either be seized by the other side, or seized by the Swiss. & The Nazis and the allies both kept their assets \hlc[lightred]{-there} \hlc[lightgreen]{+in Switzerland}. \\
         \hline \rule{0pt}{2.6ex}Why do some people vomit when they see a corpse and/or witness a homicide? & We essentially vomit at the sight of gory or bloody death as a defense mechanism. In the face of corpses or death, we are often at risk ourselves, and therefore vomit to remove possible biohazards from our system that may have been spread by the dead, as blood and gore are often good at transmitting biohazards.  \hlc[lgrey]{It also prevents us from possibly ingesting any biohazards by forcing everything out of the mouth that may have been headed for the stomach (i.e. blood)}. & \hlc[lightred]{-It also} \hlc[lightgreen]{+Vomiting} prevents us from possibly ingesting any biohazards by forcing everything out of the mouth that may have been headed for the stomach (i.e. blood). \\
         \hline \rule{0pt}{2.6ex}How does the mls all star game work? & The Major League Soccer All-Star Game is an annual soccer game held by Major League Soccer featuring select players from the league against an international club. MLS initially adopted a traditional all-star game format used by other North American sports leagues where the Eastern Conference squared off against the Western Conference. \hlc[lgrey]{This eventually evolved into the current system where the league annually invites a club from abroad to play against a league all-star team.} The MLS All-Stars hold an 8--4 record in the competition marking the season 's midpoint. \hlc[lgrey]{Players are awarded rosters spots through a combination of fan voting and selections by the appointed manager and league commissioner.} & \hlc[lightred]{-This} \hlc[lightgreen]{+The Major League Soccer All-Star Game initially adopted a traditional all-star game format used by other North American sports leagues where the Eastern Conference squared off against the Western Conference which} eventually evolved into the current system where the league annually invites a club from abroad to play against a league all-star team. Players are awarded rosters spots through a combination of fan voting and selections by the appointed manager and league commissioner. \\
         \hline
    \end{tabular}
    \caption{Decontextualization model outputs on three examples from the dataset (the summary of the original answer is highlighted in grey). Despite being out-of-domain, the decontextualization model performs reasonably well.  }
    \label{tab:decontext}
\end{table*}

\end{document}